# Hybrid-Hierarchical Fashion Graph Attention Network for Compatibility-Oriented and Personalized Outfit Recommendation

Sajjad Saed, Babak Teimourpour*

Department of Information Technology Engineering, Faculty of Industrial and Systems Engineering, Tarbiat Modares University (TMU), Tehran, Iran

{saed_s, b.teimourpour}@modares.ac.ir

*Abstract*— The rapid expansion of the fashion industry and the growing variety of products have made it increasingly challenging for users to identify compatible items on e-commerce platforms. Effective fashion recommendation systems are therefore crucial for filtering irrelevant options and suggesting suitable ones. However, simultaneously addressing outfit compatibility and personalized recommendations remains a significant challenge, as these aspects are typically treated independently in existing studies, thereby overlooking the complex interactions between items and user preferences.

This research introduces a new framework named FGAT, which leverages a hierarchical graph representation together with graph attention mechanisms to address this problem. The framework constructs a three-tier graph of users, outfits, and items, integrating visual and textual features to jointly model outfit compatibility and user preferences. By dynamically weighting node importance during representation propagation, the graph attention mechanism captures key interactions and produces precise embeddings for both user preferences and outfit compatibility.

Evaluated on the POG dataset, FGAT outperforms strong baselines such as HFGN, achieving notable improvements in accuracy, precision, hit ratio (HR), recall, and NDCG. These results demonstrate that combining multimodal visual–textual features with a hierarchical graph structure and attention mechanisms significantly enhances the effectiveness and efficiency of personalized fashion recommendation systems.

## I. INTRODUCTION

The fashion industry has grown rapidly with the rise of e-commerce and the diversification of apparel products. As online catalogs expand, users face challenges in selecting items that are both compatible and aligned with personal style preferences. To address this, fashion recommender systems have become essential tools, helping shoppers navigate vast product spaces and enhancing decision-making efficiency [1]. Beyond improving customer experience, these systems also play a critical role in reducing return rates, boosting sales, and supporting sustainable consumption by suggesting suitable items.

An outfit typically consists of a collection of fashion items (e.g., top, trousers, shoes, accessories) worn together, often tailored to a specific occasion or function. In recommendation systems, outfit compatibility refers to the degree of aesthetic and functional harmony among these items, which can be evaluated across visual, textual, stylistic, situational, and practical dimensions [2].

Compatibility modeling methods can be broadly classified into three categories: 1) pairwise models, which evaluate relationships between two items using techniques like metric learning or Siamese networks; 2) listwise models, which consider an entire outfit as a sequence or set of items; and 3) graph-based models, which capture intricate dependencies using nodes and edges to represent items and their relationships [3].

Fig. 1 illustrates an example of such a fashion graph, where compatible items form a cohesive outfit ensemble. Moreover, personalized recommendations are crucial for aligning outfit suggestions with each user's preferences, which are inferred from their interaction history and behavioral patterns. Advanced systems now incorporate collaborative filtering, attention mechanisms, or reinforcement learning to adapt to evolving user preferences.

Despite progress, a key gap remains: most recommendation systems treat outfit compatibility and personalization separately. Compatibility ensures items harmonize in style and function, while personalization aligns with individual preferences. Focusing on one without the other leads to limitations—either coherent but impersonal outfits, or personalized suggestions lacking inter-item harmony [4], [5].

Traditional recommendation techniques, such as content-based filtering and collaborative filtering, show limited effectiveness in tackling both compatibility and personalization simultaneously. Content-based methods rely on surface-level item features (e.g., color, style, texture) but struggle to model contextual relationships between fashion items [6]. On the other hand, collaborative filtering leverages user behavior data but often overlooks the visual and functional connections between outfit components [7].

Recent advancements in graph-based learning, especially Graph Neural Networks (GNNs) and Graph Attention Networks (GATs), offer promising solutions. These models represent fashion entities such as users and items as nodes in a graph and their interactions as edges, capturing complex, structured relationships. This approach allows models to learn both local dependencies (e.g., a blouse matching a skirt) and global fashion trends (e.g., community preferences), thus improving recommendation quality [8], [9].

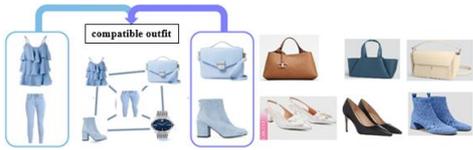

Fig. 1. A set of items forms a compatible outfit.

In this study, we introduce FGAT (Fashion Graph Attention Network), a novel Hybrid-Hierarchical Fashion recommendation framework designed to jointly address the dual challenges of outfit compatibility and user preference personalization. Unlike prior approaches that treat these aspects independently, our model integrates them into a unified, end-to-end framework. The motivation stems from the need for intelligent systems capable of understanding both the aesthetic coherence of outfits and the personal preferences of users. By integrating visual-textual modalities with graph-based relational learning, our approach overcomes the limitations of models that treat compatibility and personalization as disjointed tasks.

Our framework constructs a three-level hierarchical graph of fashion entities that contains users, outfits, and individual items. We apply Graph Neural Networks and attention mechanisms to aggregate neighborhood information at each level, considering both multimodal features and the relative importance of node pairs. This process generates refined embeddings, ultimately enabling the system to recommend outfits that are both compatible and personalized.

Our work is inspired by prior models such as HFGN [10]; however, there are two key limitations in those approaches. First, HFGN overlooks the semantic richness of item descriptions and does not incorporate textual features, which are essential for capturing complementary item characteristics that may not be visually apparent. Second, HFGN treats all nodes equally during message propagation, ignoring the varying importance of nodes and embeddings in the fashion graph, which limits its ability to model nuanced user-outfit, outfit-item, and item-item interactions. In contrast, FGAT addresses these gaps by integrating multimodal item features, combining visual representations (via ResNet) with deep semantic textual embeddings (via BERT), and by introducing a multi-tier graph attention mechanism that dynamically weighs node importance across users, outfits, and items. This enables the model to learn richer representations, capture finer-grained relationships, and significantly enhance recommendation precision.

Experiments conducted on the POG dataset demonstrate that FGAT outperforms state-of-the-art baselines, including HFGN [10], NGNN [11], and FHN [12], in key metrics such as Accuracy, Precision@10, HR@10, and NDCG@10. These results highlight the effectiveness of combining graph architectures with multimodal learning and attention mechanisms for delivering intelligent, compatibility- and user preference-aware fashion recommendations.

This study builds upon prior work in graph-based recommender systems and expands the application of GNNs and GATs to the fashion domain. By addressing the shortcomings of existing models, our proposed method offers a scalable and efficient solution to real-world challenges in fashion recommendation. The rest of the paper is organized as follows: the Related Work section reviews prior studies in fashion recommendation and graph models; the Methodology section explains the FGAT framework in detail; the Experiments section presents evaluation, discussion, and analysis of the results; and the Conclusion summarizes our findings and outlines future directions.

## II. RELATED WORK

In the following subsections, we review the key research areas that inform this study, including graph representation learning, fashion recommendation systems, and the integration of image and text processing in graph-based architectures. We also explore GNN approaches for outfit compatibility and personalized recommendation in fashion.

### A. Graph Representation Learning

Graph representation learning provides a powerful framework for modeling complex relationships among fashion entities such as users, items, and their interactions, which are essential for outfit recommendation tasks. A graph $G=(V,E)$ consists of nodes $V$ (e.g., items, users) and edges $E$ (e.g., compatibility or purchase), naturally representing the relational structure in fashion recommender systems [6], [13]. Graph theory offers foundational tools such as adjacency matrices, centrality, and traversal algorithms to analyze these relationships, where pairwise edges can capture compatibility and clusters may reveal communities with shared style preferences [14], [15].

Early methods applied shallow embedding techniques like DeepWalk [16] and GraRep [8], which learn low-dimensional node representations by preserving local or global graph structures. However, these approaches struggled with heterogeneous graphs and dynamic behaviors. More advanced methods, such as GraphSAGE [6] and HOPE [17], introduced sampling and inductive learning for scalability, while graph alignment approaches like GrAR [18] addressed structural roles across heterogeneous fashion graphs.

Recent advances in GNNs have significantly enhanced representation learning by aggregating neighbor information through neural layers [19], [20]. GATs further refine this process with attention and emphasizes more relevant relationships, such as compatibility between items in an outfit [21]. In fashion applications, nodes often represent items with multimodal attributes (e.g., images, text), and edges encode compatibility or user–item interactions, enabling graph-based models to predict user preferences and generate coherent outfit recommendations [11], [22].

### B. Fashion Recommender Systems

Traditional recommender methods such as content-based and collaborative filtering face limitations: content-based systems fail to ensure outfit-level coherence, while collaborative filtering often ignores functional harmony between items [1], [23]. Deep learning and hybrid models have improved feature learning, but they frequently treat compatibility and personalization separately—leading to mismatched or less personalized recommendations.

To overcome these issues, recent studies adopt graph-based approaches, where fashion items are modeled as nodes and relations such as co-purchase or user–item interactions as edges [24], [25]. These methods capture higher-order and non-linear dependencies, allowing more nuanced modeling of fashion relationships [26]. GNNs propagate information across connected nodes to encode both user preferences and compatibility signals, while GATs refine this process by prioritizing the most relevant neighbors [10], [27]. Moreover, hierarchical neural architectures, such as those proposed by Chen et al. [28], provide structured and abstract modeling of multi-level relationships in fashion graphs.

C. *Image Processing in Fashion*

Image processing plays a central role in fashion recommendation systems by extracting visual features, such as color, texture, pattern, and shape, that determine outfit compatibility [29]. Traditional techniques like edge detection and color histograms offered limited insights, but deep learning, particularly Convolutional Neural Networks (CNNs) and Multi-CNNs, revolutionized feature extraction, capturing higher-level aesthetic properties [31], [32]. Following the success of Krizhevsky et al.'s ImageNet model [33], CNN-based systems widely supported tasks such as outfit matching and similarity scoring, with VGG and ResNet generating robust embeddings. Advanced approaches like Siamese/Triplet Networks improved similarity learning, R-CNNs enabled fine-grained garment analysis, and GANs addressed data scarcity by synthesizing clothing images [34]. Major platforms such as Amazon and Zalando also leveraged visual features for personalized style-matching recommendations [35].

Between 2016 and 2020, multimodal models emerged, integrating visual, textual, and behavioral data. Fusion and co-attention techniques enhanced personalization, while models such as HMGN [37] advanced outfit-level compatibility prediction [36]. More recently, research has shifted toward hyper-personalization and sustainability. Transformer-based vision–language models [7] enable nuanced interpretation of images and text, tailoring recommendations to individual style [39], body type [40], and even mood [41]. At the same time, eco-conscious systems now identify sustainable materials, vintage fashion, and second-hand clothing [38], [42].

D. *Text Processing in Fashion*

Text processing is crucial in fashion recommendation systems for extracting functional and contextual information—such as material, brand, occasion, and user sentiment—that complement visual features and enhance recommendation quality. Unlike image-based methods, text processing captures semantic details from product descriptions, reviews, and metadata [43].

Traditional approaches like bag-of-words and TF-IDF provided shallow insights, but deep learning introduced richer semantic modeling. Word embeddings (Word2Vec, GloVe) capture semantic similarities (e.g., "shirt"–"trousers"), improving the interpretation of product descriptions [44]. Sequential models such as RNNs and LSTMs further advanced the field by handling context in reviews and long descriptions, while transformers, particularly BERT, leveraged attention mechanisms to focus on key attributes, achieving stronger contextual understanding [45], [46].

Multimodal systems integrate textual and visual data for more comprehensive representations; for example, FashionVII combines descriptions and images to enhance attribute recognition [7]. Text also supports personalization through reviews and search queries [47]. Sentiment analysis is widely applied to infer user satisfaction, with classifiers identifying positive or negative opinions about specific items [48]. Topic modeling methods like LDA further extract themes (e.g., "formal wear," "athleisure") to aid categorization and tailored recommendations [49].

Despite progress, challenges remain in effectively fusing textual and visual data, and in extracting user-specific preferences from sparse or noisy reviews [1].

E. *Graph Neural Networks Approach for Outfit Compatibility*

Early works applied graph structures to model outfit compatibility. He et al. introduced a bipartite graph for user–item interactions [50], while Cucurull et al. incorporated visual signals for compatibility prediction [51]. Style2Vec by Lee et al. learned item embeddings from outfit contexts [52], and Veit et al. modeled aesthetic compatibility via co-occurrence patterns [2]. More recently, OutfitTransformer leveraged attention and task-specific tokens to improve outfit retrieval and compatibility [53].

Deep graph-based methods combine neural learning with structured graphs to address fashion complexity. Chen et al. applied deep learning for personalized outfit generation on iFashion [54], while Vasileva et al. proposed type-aware embeddings integrating visual and textual features [39]. Guan et al. enhanced coherence through multimodal compatibility modeling and semi-supervised learning [3], [55].

GNNs further advanced outfit modeling by propagating information across direct and indirect item relations. Cui et al.'s NGNN generated cohesive outfits via node propagation [11], while Abu-El-Haija et al.'s N-GCN and Hu et al.'s models addressed semi-supervised and unsupervised learning [56], [57]. Guan et al. also proposed bidirectional hash-graph methods for efficient retrieval [58]. More recent advances include Vision Transformer embeddings combined with hypergraph networks for compatibility prediction [59], and strategies addressing data imbalance through novel pooling and negative sampling [60].

F. *Textual and Visual Models*

Integrating textual and visual modalities improves fashion item understanding. He and McAuley's VBPR model incorporated visual cues into Bayesian ranking [61]. Sagar et al. introduced interpretability in personalized recommendations with PAI-BPR [62].

Dong et al. proposed a multimodal try-on framework combining visual, textual, and behavioral data [63]. Song et al.'s modality-aware learning dynamically weighted features based on context [64]. These models demonstrate the value of multimodal fusion for better outfit compatibility.

G. *Personalized Recommendation in Fashion*

Personalization in fashion recommendation has been advanced through graph-based methods. Early approaches include tensor factorization for personalization [65] and adaptations of GCNs to incorporate user data [5]. Lu et al. introduced binary code learning for efficient personalized recommendations [12], while Song et al.'s GP-BPR modeled personalized compatibility by treating items as graph nodes [66]. Li et al. proposed HFGN, a hierarchical graph framework that effectively captured user–outfit–item relationships [10].

Subsequent works enhanced modeling of complex relations: Guan et al. developed a meta-path guided heterogeneous graph framework [67], Vivek et al. proposed OGN for style-profile-based compatibility [68], and Wang et al. leveraged multimodal conditional preferences [69]. Liu et al. introduced a self-adjusting GNN using category co-occurrence [27], while Xu et al.'s HMGL-OCM combined item- and state-level graphs to improve real-world performance [70]. Other approaches integrated biometric and seasonal color features [71] or efficient hybrid models like PFRS using EfficientNet and KNN [72].

Recent advances include transformer-based personalization from purchase history [73], hierarchical latent modeling with DGHNet [74], co-attention and meta-path learning in GPA-BPR [75], and knowledge-based methods like StePO-Rec for transparent recommendations [76]. Further innovations include BiHGH with bidirectional convolution in quadripartite graphs [77] and feature-graph models addressing cold-start issues [78].

Building on these efforts, the proposed FGAT model integrates hierarchical GNNs, multimodal fashion data, and attention mechanisms to jointly capture compatibility and personalization, outperforming baselines such as HFGN [10], NGNN [11], and FHN [12] on the POG dataset.

## III. METHOD

The proposed FGAT framework is designed to simultaneously address two key challenges in fashion recommendation: outfit compatibility modeling and personalized user preferences. Inspired by prior hierarchical graph architectures such as HFGN [10], our method introduces significant enhancements through the integration of multimodal (visual-textual) features, graph attention mechanisms and a three-level hierarchical graph structure that jointly encodes users, outfits and items.

Fig. 2 illustrates the general flow of our proposed method. As shown, the framework consists of three main components:

- We initialize embeddings for users and outfits using IDs, and for items using visual-textual features.
- Updating embeddings via GATs using first-order paths. It should be noted that in this method we use first-order paths (outfit–item and user–outfit), while higher-order paths (such as user–outfit–item) are left for future work.
- We predict personalized outfit compatibility scores and recommend personalized outfits.

### A. Data Collection

To evaluate the effectiveness of the proposed FGAT framework, we used the POG dataset [54]. This dataset contains rich user interactions, outfit compositions, and item-level attributes, which are essential for modeling both compatibility and personalization in fashion. The dataset includes:

- User data which records the user's previous purchases and interactions on outfits.
- Outfit data which each outfit is composed of multiple fashion items (e.g., tops, bottoms, shoes).
- Item data which provides the item context details, including: category, title (in Chinese), and download link for the item image. In total, we have 61 different categories such as: coat, pants, shirt, necklace, etc. for fashion items.

For evaluating model in recommendation task, 80% of each user's interactions are used for training, 10% of that training set is used for validation, and the remaining 20% is used for testing. Also for compatibility task, the training set includes positive outfit samples from the recommendation dataset to learn item compatibility. Additionally, 1,647 unused/ unappeared items in training set, are selected as negative samples for the test set. Dataset statistics are provided in Table I. and Table II.

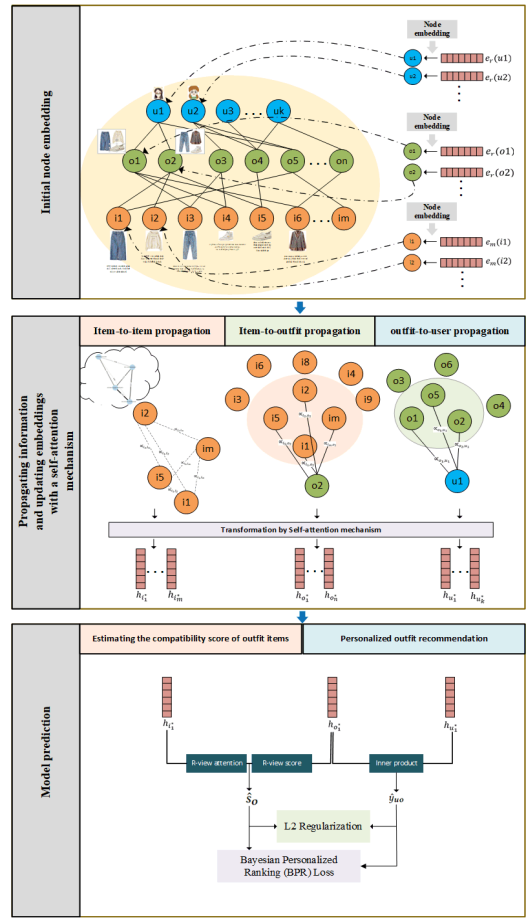

Fig. 2. The proposed FGAT scheme comprises three key components: (1) initial node embedding, (2) information propagation and embedding updates via self-attention mechanism, and (3) model prediction.

TABLE I. DATASET STATISTICS FOR THE PERSONALIZED OUTFIT RECOMMENDATION TASK.

| *users* | outfits | items | interactions |
|---|---|---|---|
| *38,415* | 9,373 | 19,175 | 274,542 |

TABLE II. DATASET STATISTICS FOR ESTIMATING THE OUTFIT COMPATIBILITY SCORE TASK.

| *Dataset* | outfits | items |
|---|---|---|
| *Training* | 9,373 | 19,175 |
| *Test* | 1,647 | 3,126 |

## B. Graph Construction

After preparing the dataset, a three-level heterogeneous fashion graph is built with users, outfits, and items as nodes (Fig. 3). Edges from level 1 to level 2 represent user interactions with outfits, and edges from level 2 to level 3 represent the items that form a compatible outfit. According to the statistics in Table I., the graph contains 38,415 user nodes, 9,373 outfits nodes and 19,175 item nodes — totaling 66,963 nodes and 274,542 undirected edges.

## C. Initial Node Embedding

To represent hidden features, each user/outfit ID is mapped to a vector representation as its initial embedding. Specifically, user $u$ and outfit $o$, are embedded as $u \in R^d$ and $o \in R^d$, with embedding dimension $d = 64$. For clarity, the notations used throughout this paper are summarized in Table III.

- Visual Feature Extraction of Item:

Each fashion item $i$ has visual features $x_v(i)$. Since items belong to different fashion categories, category-aware encoders are used to extract 32-dimensional visual embeddings. The ResNet-152 pre-trained model is used to extract these features (1):

$$e_v(i) = Resnet152(x_v(i)) \qquad (1)$$

The embedding resulting from the extraction of item visual features by mapping a two-layer MLP ($f_c$) is shown below (2):

$$\hat{e}_v(i) = f_c(e_v(i)) \qquad (2)$$

- Textual Feature Extraction of Item:

In addition to visuals, textual features such as descriptions or titles offer valuable context that helps better understand items and their relationships to users and outfits. For example, as shown in Fig. 4, an item image can provide visual cues such as color or pattern, while the description/title text may reveal complementary information like material (e.g., leather) or gender-specific use (e.g., women's shoes). Combining these modalities provides a more comprehensive view of the item. Since the text is in Chinese, a pre-trained Chinese BERT model is used to extract 32-dimensional textual embeddings (3), (4).

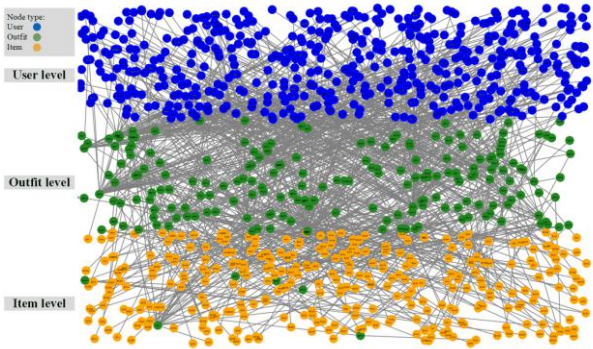

Fig. 3. A three-level fashion graph of users, outfits, and items.

TABLE III. SUMMARY OF THE MAIN NOTATIONS.

| Notation | Explanation |
|---|---|
| $u$ | A user in the system. |
| $o$ | An outfit in the system. |
| $i, j$ | Indices representing items in outfits. |
| $c_i, c_j$ | The category of item $i$ and item $j$, respectively. |
| $R^d$ | Embedding d-dimensional real vector space. |
| $x_v(i)$ | Visual features of item i. |
| $x_t(i)$ | Textual features of item i. |
| $\hat{e}_v(i)$ | Visual embedding of item i. |
| $\tilde{e}_t(i)$ | Textual embedding of item i. |
| $e_m(i)$ | Final embedding of item i. |
| $f_c$ | fully-connected layer |
| $h_i$ | Initial embedding of item $i$. |
| $h_{i*}$ | Updated embedding of item $i$. |
| $h_o$ | Initial embedding of outfit o. |
| $h_{o*}$ | Updated embedding of outfit o. |
| $h_u$ | Initial embedding of user u. |
| $H_{u*}$ | Updated embedding of user u. |
| $W, W_2, W_3$ | Learnable transformation matrices ($\in R^{d \times d}$) |
| $a$ | Learnable attention weight vector (a $\in R^{2d}$) |
| $\alpha_{i,j}$ | Attention weight from item $j$ to item $i$. |
| $e_{i,j}$ | Attention coefficient between item $i$ and item $j$. |
| $N_i$ | Neighborhood set of item $i$. |
| $w(c_i, c_j)$ | Co-occurrence weight between categories $c_i$ and $c_j$. |
| LeakyReLU(.) | LeakyReLU Activation function. |
| $\sigma(\cdot)$ | Sigmoid activation function. |
| tanh(.) | hyperbolic tangent activation function |
| softmax(.) | Softmax activation function. |
| $\odot$ | Element-wise product. |
| ‖·‖, [] | Concatenation operator. |
| $\Theta$ | Set of model parameters. |
| $L_{comp}$ | Outfit compatibility score prediction loss. |
| $L_{rec}$ | Personalized outfit recommendation loss. |
| $O_{em}$ | Embedding matrix of an outfit ($O_{em} \in R^{n \times d}$). |
| $W_4$ | Learnable weight matrix for R-view attention($W_4 \in R^{R \times v}$). |
| $W_5$ | Learnable weight matrix for R-view attention($W_5 \in R^{v \times d}$). |
| $W_6, W_7$ | Learnable weight matrix for R-view compatibility score. |
| $R$ | Number of semantic views. |
| $H$ | Training set for personalized outfit recommendation. |
| $H'$ | Training set for outfit compatibility score prediction. |
| $A_{rm}$ | R-view attention matrix |
| $C_{rm}$ | R-view compatibility score matrix |
| $a_r$ | r-th row of the attention matrix $A_{rm}$. |
| $c_r$ | r-th row of the compatibility matrix $C_{rm}$. |
| $k$ | the number of top recommendations considered for evaluation. |

$$e_t(i) = BERT(x_t(i)) \tag{3}$$

$$\tilde{e}_t(i) = f_{cls}(e_t(i)) \tag{4}$$

Here, $f_{cls}(.)$ represents the embedding space mapping. Specifically, we use the average hidden states corresponding to the special fully-connected token at the beginning of the input sequence, i.e., [CLS], from the last two layers of BERT as the representation of textual descriptions.

BERT captures bidirectional deep contextual of a word in the text and semantic meaning, making it more effective than older models (e.g., Word2Vec, LSTM) for modeling complex item-item relationships.

- Final Item Embedding:

Visual and textual embeddings are concatenated and passed through a learnable fully-connected (FC) layer to project into the same 64-dimensional space as users and outfits (5):

$$e_m(i) = f_c([\hat{e}_v(i), \tilde{e}_t(i)]) \tag{5}$$

Fig. 4 shows the initial embedding of the user, outfit, and item nodes.

### D. propagating information and updating embeddings via self-attention mechanism

Recent research in GNNs ([3],[5],[10],[11],[55],[67], [68]) has shown that propagating information in graph structures can extract valuable multi-hop neighbor information and encode higher-order relationships into node representations. Inspired by these works, we implement a hierarchical graph network with a self-attention mechanism to perform embedding propagation over the fashion graph, leading to improved embeddings. It captures complex relationships like item compatibility and user preferences through three stages:

- Item-to-item propagation: Enhances item embeddings by modeling their compatibility.
- Item-to-outfit propagation: Integrates item-level meanings into outfit embeddings.
- outfit-to-user propagation: Combines users' historical purchases to form user representations.

o **Item-to-item propagation**

Items are at the lowest level of the fashion graph and represent individual visual and textual features, as well as compatibility relationships. For example, connections like $\{i_4, i_7\} \rightarrow o_1$ not only show that items $i_4$ and $i_7$ belong to outfit $o_1$ but also indicate that these items are compatible. Such compatibility implies that compatible items should propagate more information than unrelated ones. To model this explicitly, we first construct a category graph for all outfits and then a subgraph of items for each outfit.

❖ Category Graph Construction

Each item belongs to a specific category (e.g., shirt, pants, sandals). A weighted category graph is built, where edge weights are based on how often categories co-occur in outfits. For example, bags are paired more frequently with jackets than sandals, so the bag–jacket edge has a higher weight. This co-occurrence weight is obtained according to Equation (6):

$$w(c_i, c_j) = \frac{co(c_i, c_j) / o(c_j)}{\sum_{c_k} co(c_i, c_k) / o(c_k)} \tag{6}$$

Where $co(c_i, c_j)$ is the number of times categories $c_i$ and $c_j$ (e.g., shirt and pants) appeared together (co-occurrence) in outfits, and $o(c_j)$ is the number of times a particular category $c_j$ appeared in all the outfits. You can see category graph in Fig. 5, where the nodes, categories, and edges represent the weighted co-occurrence of categories in the outfits.

Any two categories that appear together more frequently have a stronger relationship (i.e., higher weight) in this graph. For example, Fig. 6 shows the top 5 category pairs that co-occur most frequently in the outfits.

❖ Item-item Subgraph Per Outfit

Using the category graph, a subgraph is created for each outfit. Only the items within that specific outfit are included, and edge weights are inherited from the category graph (Equation (6)). This reflects the coarse-grained compatibility of items in an outfit. Fig. 7 shows a representation of the item subgraph for five sample outfits.

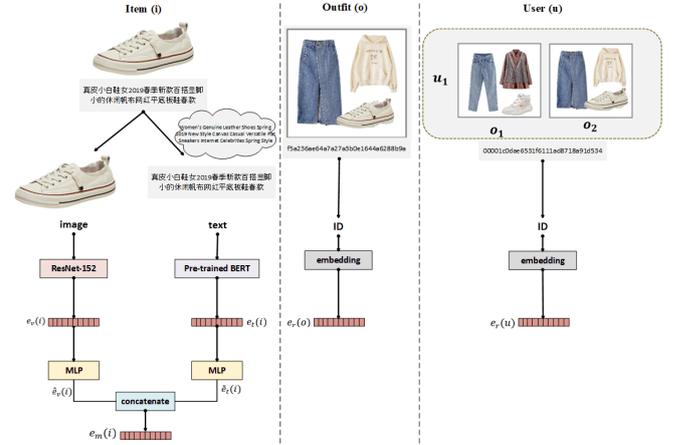

Fig. 4. Initial embedding of user, outfit and item.

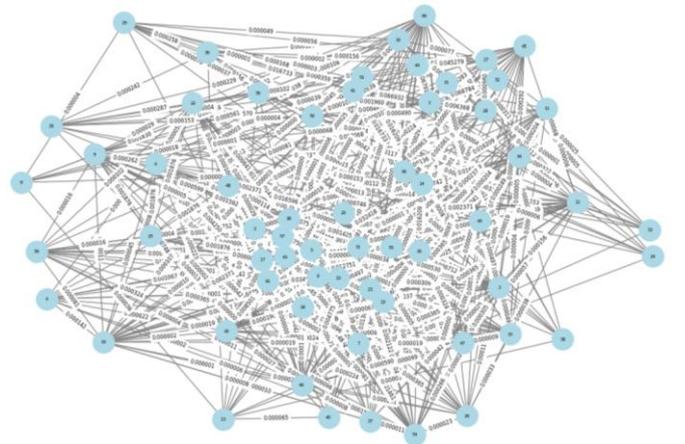

Fig. 5. An undirected category-category graph where weighted edges represent the co-occurrence of categories.

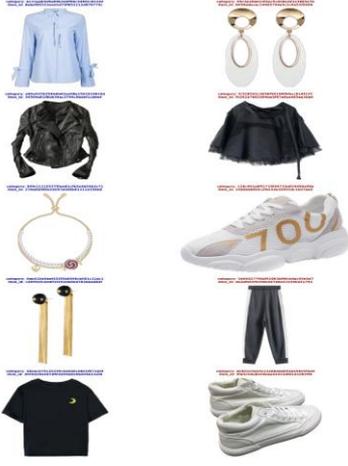

Fig. 6. Top 5 pairs of categories that co-occur most frequently in the outfits.

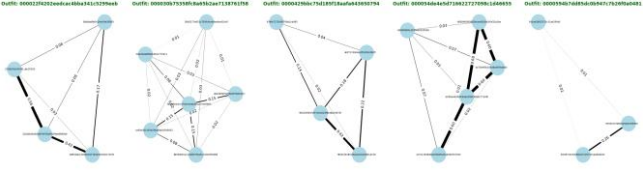

Fig. 7. Item subgraph of 5 sample outfits. Nodes represent items, and edge weights indicate the co-occurrence frequency of item categories that appear together in the same outfit.

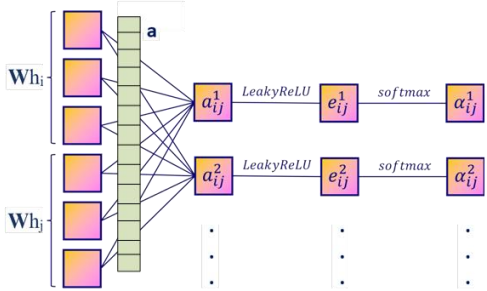

Fig. 8. Multi-head attention.

❖ Aggregation of item neighborhood information with self-attention mechanism and update embedding

Using a self-attention mechanism, each item aggregates compatibility signals from its neighbors. The calculation of the attention coefficient $e_{i,j}$ using the learnable weight vector **a** and the transformation matrix $W$ is as follows:

$$e_{i,j} = LeakyRelu\ (a^T[Wh_i || Wh_j]) \quad (7)$$

where $W \in R^{d \times d}$ maps the features $h_i$ and $h_j$ to a new space that is initialized by the weight value $w(c_i, c_j)$ calculated in the previous steps (Equation (6)), where $c_i$ and $c_j$ are the categories of items i and j, respectively. In this way, the $w(c_i, c_j)$ also reflects the compatibility between item categories. The symbol || denotes the concatenation of the embeddings of items i and j. The vector $a \in R^{2d}$ learnable weight vector used to compute the attention score from the concatenated embeddings $h_i$ and $h_j$, and LeakyReLU(.) is the activation function applied to control small gradients.

The attention weight $\propto_{i,j}$ between two items i and j is calculated by normalizing the attention coefficient $e_{i,j}$ as follows, indicating how important item j is for updating the information of item i:

$$\propto_{i,j} = \frac{\exp(e_{i,j})}{\sum_{k \in N_i} \exp(e_{i,k})} \quad (8)$$

Here, exp denotes the exponential function used to normalize attention values, and $\sum_{k \in N_i}$ is the sum of all attention scores that item i receives from its neighbors, ensuring that the total adds up to 1. A larger value of $\propto_{i,j}$ implies that item j plays a more significant role in updating the representation of item i.

Typically, graph-based models with attention mechanisms employ multi-head attention, allowing the model to attend to different aspects of neighbors simultaneously and extract more complex features. This structure improves the model's stability and learning capacity by reducing fluctuations. Moreover, by averaging across multiple heads, the model achieves better generalizability and interpretability. In this work, we also use 4-head multi-head attention, whose schematic is shown in Figure 8. The message passed from neighbor $j$ to item $i$ is:

$$m_{j \to i} = LeakyReLU(\propto_{i,j} W_1(h_i \odot h_j)) \quad (9)$$

Here, $\odot$ denotes element-wise product, used to incorporate the visual and textual compatibility of items i and j, encouraging more compatible items to contribute more to the message. Each item's embedding is updated by aggregating messages from its neighbors using the following:

$$h_{i^*} = h_i + LeakyReLU(\sum_{j \in N_i} \propto_{i,j} W_1(h_i \odot h_j)) \quad (10)$$

Where $h_{i^*}$ is the updated embedding of item i. The initial embedding $h_i$ was obtained in the previous steps according to Equation (5), denoted as $e_m(i)$. This step ensures that compatible neighbors contribute more to the updated representation, while unrelated items contribute less. Additional layers can be added in future work to capture higher-order relations.

o **Item-to-outfit propagation**

Intuitively, an outfit can be described based on its component items. But do all items contribute equally to defining an outfit? Not necessarily. Some items—such as jackets or shirts—have a greater influence on determining the overall style, while others—like belts or socks—play a less significant role.

Solution: We use attention mechanisms to enable the model to identify which items contribute more significantly to the outfit's overall composition.

❖ Item- Outfit Heterogeneous Graph

We construct a heterogeneous graph consisting of two types of nodes: items and outfits. Edges in this graph represent the relationship between each item and its corresponding outfit.

❖ Aggregation of Outfit Neighborhood Information Using Self-Attention and Embedding Update

Focusing on a specific outfit, we refine its representation by aggregating the embeddings of its component items. The message passed from item i to outfit o is defined as following

Equation(11). $h_{i*}$, is the updated embedding of item i (see Equation (9)), and $W_2$ is a learnable transformation matrix.:

$$m_{i \to o} = \text{LeakyReLU}(\alpha_{i,o} W_2 h_{i*}) \quad (11)$$

The attention weight $\alpha_{i,o}$ indicates the importance of item i for representing outfit o, and is defined as follows:

$$\alpha_{i,o} = \frac{\exp(e_{i,o})}{\sum_{j \in N_o} \exp(e_{i,o})} \quad (12)$$

And The calculation of the attention coefficient $e_{i,o}$ using the learnable weight vector **a** and the transformation matrix W is as follows:

$$e_{i,o} = \text{LeakyRelu}(a^T[Wh_{i*}||Wh_o]) \quad (13)$$

As in item-level aggregation, the final representation of the outfit, $h_{o*}$ is generated by combining its initial embedding $h_o$ with the attention-weighted updated embeddings of its associated items ($h_{i*}$):

$$h_{o*} = h_o + \text{LeakyRelu}(\sum_{i \in N_o} \alpha_{i,o} W_2 h_{i*}) \quad 14$$

Unlike traditional methods that rely solely on visual features, this approach also incorporates item compatibility scores. The attention mechanism enables the model to weigh each item's contribution to the outfit, producing embeddings that are more accurate, comprehensive, and style-aware.

- o **Outfit-to-user propagation**

A user's history reflects their fashion preferences. For instance, by analyzing past outfits (e.g., $o_1$ and $o_2$) selected by user $u_1$, we can infer their style. The connection $o_5 \to \{u_1, u_2\}$ also suggests behavioral similarity between users. Additionally, user histories offer collaborative filtering signals — users with similar behaviors often share similar outfit preferences.

To capture this, we improve user ID embeddings by incorporating the representations of the outfits they've interacted with, using a heterogeneous graph structure.

❖ Outfit- user Heterogeneous Graph

We also build a heterogeneous graph that contains two types of nodes: users and outfits, and edges that indicate interactions (e.g., purchases or likes).

❖ Aggregation of User Neighborhood Information Using Self-Attention and Embedding Update

For a target user u, we focus on their interacted outfits ($N_{(u)}$) and extract useful signals from each outfit is defined as follows. $h_{o*}$ is updated embedding of outfit o and $W_3$ is a learnable matrix for transformation:

$$m_{o \to u} = \text{LeakyReLU}(\alpha_{o,u} W_3 h_{o*}) \quad 15$$

$\alpha_{o,u}$ : attention weight showing how relevant o is to user u, computed as:

$$\alpha_{o,u} = \frac{\exp(e_{o,u})}{\sum_{j \in N_u} \exp(e_{o,u})} \quad 16$$

$$e_{o,u} = \text{LeakyRelu}(a^T[Wh_{o*}||Wh_u]) \quad 17$$

This allows the model to learn which outfits most influence a user's style representation. The final user representation ($h_{u*}$) is computed as:

$$h_{u*} = h_u + \text{LeakyReLU}(\sum_{o \in N_u} \alpha_{o,u} W_3 h_{o*}) \quad 18$$

This representation combines ID-based features (inherent user characteristics) and style-aware features (based on outfit history). Unlike earlier methods that used only user IDs, this approach yields richer, more accurate user representations. By propagating information through the hierarchical fashion graph, the model effectively integrates complex relationships across items, outfits, and users to improve learning and personalization.

IV. EXPERIMENT

The POG dataset, which was introduced earlier in the "Methodology - A. Data Collection" section, was used for the empirical evaluation of the proposed FGAT model. The fashion graph, which contains 66,963 nodes and 274,542 undirected edges, was visualized using the Gephi application and is shown in Fig. 9.

A. Attention Weight Initialization

A key innovation is the initialization of item-item attention weights based on category co-occurrence in outfits. More frequent co-occurrences between categories (e.g., shirts & pants) imply stronger compatibility and thus higher initial attention weights. These weights help the model prioritize compatible items during information propagation. A heatmap and scatter diagram (Fig. 10) visualize category pair frequencies and attention scores—for example, the strong compatibility between floral shirts and matching shoes/bags (Fig. 11).

As shown in Fig. 10, the two categories appeared together nearly 3,500 times in different outfits, indicating a strong and consistent relationship between them. A large number of category pairs appeared together in the range of about 1,500 to 3,000 times, while the remaining pairs appeared together less frequently in outfits.

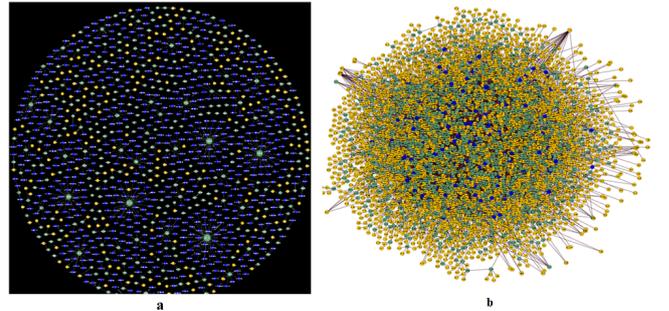

Fig. 9. Visualization of a three-level fashion graph using the Gephi application: a) Outfits purchased by a large number of users are represented with many edges. b) Items that appear in multiple outfits are also represented with many edges.

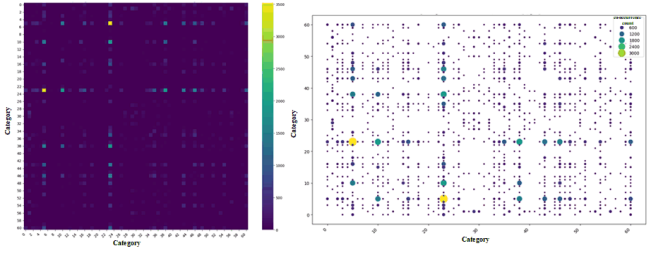

Fig. 10. Heatmap and scatter diagram of category co-occurrence in the same outfits.

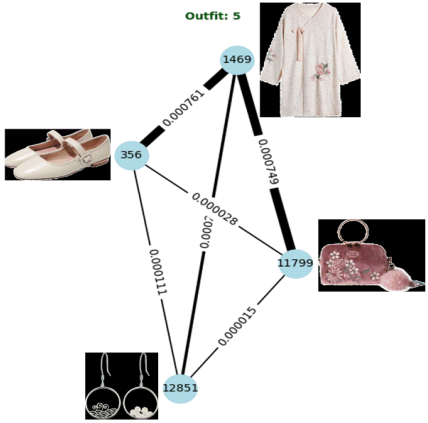

Fig. 11. An outfit showing different attention weights for a pair of items.

The self-attention mechanism in the outfit items subgraph stage adjusts attention weights based on co-occurrence relationships, such that more related items receive more weight. Category co-occurrence helps the model prioritize items that are most likely to be compatible in personalized recommendations. For example, if a user has shown interest in a shirt, the model is more likely to recommend pants or shoes that frequently co-occur with the shirt.

Fig. 11 shows an example of co-occurrence weights of items according to their categories for outfit sample 5. As shown, the edge between the cream floral shirt and the cream shoes, as well as the floral bag that matches the shirt, receives a higher weight.

In the information propagation stage between items, these attention weights serve as signals to update the item embeddings. Items that belong to categories with high similarity exert greater influence on each other's embeddings. For example, by incorporating these attention weights during embedding updates, the model better recognizes the association between a cream-colored floral shirt and a floral bag of a similar pattern but different color.

By recognizing patterns of similarity, the model can make suggestions that align not only with the user's preferences but also with fashion standards, thus increasing the accuracy of matching. With this strategy, the model is more likely to prioritize the bag as a compatible item over earrings when suggesting items for an outfit that includes the shirt.

### B. Model Implementation settings

To implement the proposed model, experiments were conducted in the Google Colab environment using graph libraries such as NetworkX, Matplotlib, and PyTorch Geometric. The dataset was split as follows: 80% for training, 10% for validation, and 10% for testing. We set the embedding dimension to 64, batch size to 512, learning rate to 0.001, dropout rates to 0.2 and 0.3, enabled batch normalization, and applied L2 regularization to prevent overfitting. The Adam optimization algorithm was used.

### C. Model Implementation results

We implemented a joint learning scheme that enables the simultaneous matching of outfit items and personalized outfit recommendation. The model training results across 100 epochs for different metrics are shown in Fig. 12.

To optimize the model, we used the Bayesian Personalized Ranking (BPR) [79] loss function for both the personalized outfit recommendation task and the prediction of outfit compatibility scores. BPR assumes that existing interactions have higher prediction scores than non-existent interactions. The objective functions for the two tasks are:

$$L_{rec} = \min_{\Theta} \sum_{(u,o,o') \in \mathcal{H}} -\ln \sigma(\hat{y}_{uo} - \hat{y}_{uo'}) \quad 19$$

$$L_{com} = \min_{\Theta} \sum_{(o,o') \in \mathcal{H}'} -\ln \sigma(\hat{s}_o - \hat{s}_{o'}) \quad 20$$

Where $\mathcal{H}=\{(u,o,o')\}$ is the training set for personalized outfit recommendation, with each triple $(u,o,o')$ representing a historical interaction of user u with outfit o and a non-existent interaction with outfit o'. $\mathcal{H}'=\{(o,o')\}$ is the training set for outfit compatibility score modeling, where each pair $(o,o')$ includes an existing outfit o (i.e., a positive samples) and an non-existent outfit o' (i.e., a randomly generated negative samples); σ(.) is the sigmoid function, $\hat{y}_{uo}$ is the probability of user u purchasing outfit o (Equation (21)) , $\hat{s}_o$ is the compatibility score for the outfit o (Equation (24)) and Θ denotes the set of model parameters. $L_{rec}$ is the personalized outfit recommendation loss, and $L_{com}$ is the compatibility score prediction loss.

### D. Evaluation Metrics

To evaluate the performance of top-K recommendation, we consider the HR@K, NDCG@K, Recall@K, and Precision@K metrics (K=10). For all metrics, we report the average across all users as the final score. For the outfit compatibility score prediction task, we conducted the Fill-in-the-Blank (FLTB) task and used the accuracy and AUC metrics to evaluate the model's performance.

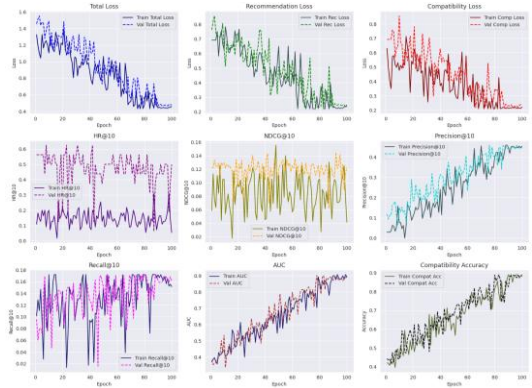

Fig. 12. Evaluation charts of various metrics during model training over 100 epochs.

*E. Discussion and Analysis of the Results*

The proposed model—by combining a graph-based structure, leveraging the visual and textual features of items, incorporating category co-occurrence weights in the graph attention mechanism, and propagating information across three levels—achieved promising results, which are analyzed and discussed below.

The **recommendation loss** illustrates the reduction in model error for the personalized outfit recommendation task. Both the training (black line) and validation (dotted green line) curves generally show a downward trend, indicating effective model learning over the epochs. Some fluctuations are expected due to the data's variability and the model's complexity. However, between epochs 75 and 100, both curves demonstrate good stability. By the end of training, the loss value dropped below 0.25, indicating high precision in the model's recommendations. This suggests that the model has effectively learned user preferences and progressively reduced prediction errors.

The **compatibility loss** measures the model's performance in predicting outfit compatibility scores (i.e., whether items in an outfit set match well). Both the training and validation curves show a consistent downward trend. The decrease in error from around 0.7 to below 0.25 toward the end suggests that the model has successfully learned the relationships between items in an outfit. The close alignment of the training and validation curves indicates minimal overfitting. Consequently, the model effectively captures item compatibility and significantly reduces the likelihood of incorrect matches.

The **total loss** shows the model's overall performance across both tasks. A steady and significant decrease is observed in both the training and validation curves. The total loss value declined from approximately 1.5 to around 0.5, reflecting strong convergence and simultaneous improvement across different components of the model. Although some fluctuations were observed in the middle stages of training, the model exhibited solid stability in the final epochs.

**HR@K:** This metric evaluates the recommender system's precise in retrieving outfits that users actually liked. If the correct outfit (i.e., one the user interacted with) appears among the top K suggestions, HR is 1; otherwise, it is 0. This value is averaged across all users. It indicates whether the system can include at least one relevant item in its top-K recommendations.

The results for HR@10 revealed some instability in the validation set, with values fluctuating between 0.4025 and 0.6250. Sharp drops occurred at epochs 16, 20, 21, 45, and 86. Interestingly, lower HR@10 values in the training set compared to the validation set may point to overfitting, optimization challenges, or issues related to random negative sampling. Ultimately, the model achieved an HR@10 of 0.4286 on the test set—outperforming HFGN—and reached its best training performance at epoch 95, with an HR@10 of 0.4319.

**Recall@K:** This is the ratio of correctly recommended outfits by the model to the total number of relevant outfits. It measures how many relevant outfits the model successfully identifies.

The results for this metric were better than HR@10, with smaller differences between training and validation. In the final epochs, Recall@10 stabilized between 0.16 and 0.18. On the test set, Recall@10 was 0.1580, and the best training performance was at epoch 90 with Recall@10 = 0.18.

**Precision@K:** This is the ratio of correct (relevant) items among the top K recommendations, indicating the precision of the recommender system. It is calculated as the number of correct items among recommendations divided by K.

The results show an upward trend with acceptable fluctuations and a small gap between training and validation, indicating good model performance. The model's attention mechanism incorporates category co-occurrence weights and propagates embeddings to improve precision in compatibility and personalization.

At epoch 86, the best training precision was 0.4525, validation reached 0.4448, and on the test data, Precision@10 was 0.4424.

**NDCG@K:** (Normalized Discounted Cumulative Gain) is a widely used metric to evaluate the quality of a ranked list by considering the relevance and positions of items in the Top-K recommendations. Higher-ranked relevant items contribute more to the score. This metric evaluates how well the model ranks good outfits at the top.

On the test data, NDCG@10 was 0.1340, an improvement over baseline models such as HFGN (0.1241). The best training NDCG@10 value of 0.1412 was obtained at epoch 89.

**AUC:** The probability that a positive item scores higher than a negative item. It measures the model's ability to discriminate between related and unrelated items. Calculated based on the ROC curve, which plots the True Positive Rate (TPR) against the False Positive Rate (FPR).

The model's AUC on the test set was 0.8974, with the best training performance of 0.9014 at epoch 90. Initial fluctuations in AUC were observed during training, but stability improved after optimizing hyperparameters such as learning rate and tuning. Propagating information in a three-level graph and combining visual and textual features helped produce more precise embeddings for items, outfits, and users, enhancing AUC.

**Accuracy:** Represents the ratio of correct predictions to total predictions, calculated as: number of correct predictions divided by total number of predictions.

In recommender systems, especially for outfit matching and predicting outfit compatibility scores, accuracy evaluates the model's ability to identify compatible items (e.g., selecting proper items to complete an outfit in a Fill-in-the-Blank task). This metric is useful when compatible and incompatible classes are balanced.

In the proposed model, accuracy was used for the outfit compatibility prediction task, where the model selects the correct item from positive and negative options. The model achieved an accuracy of 0.8956 on the test set, improving over baselines such as HFGN (0.8797). The best training accuracy was 0.8991 at epoch 92.

This improved performance is attributed to the attention mechanism, which assigns different importance weights to nodes during information propagation, instead of uniform weights for neighbors, as well as the concatenation of visual and textual features of items.

## F. Model Comparison

Table IV. compares the experimental results of the proposed model with baseline models for the personalized outfit recommendation task. The last column shows the significant improvement of the proposed method over the baselines in HR@10 and Precision@10. The proposed model achieves the highest performance in these two metrics, showing it is better at ranking correct outfits within the top-10 recommendations and suggesting more prices results. For Recall@10, the model ranks second-best, meaning it retrieves slightly fewer relevant items compared to the best baseline, but still competitive. On NDCG@10 (which considers ranking positions), the model outperforms HFGN (from 0.1241 to 0.1340), confirming better ranking quality and prioritization of relevant outfits.

Table V. compares the accuracy and AUC results of the proposed model with baseline models for the task of predicting outfit compatibility scores and matching items in the FLTB task. The proposed model consistently improves over baselines, confirming it learns compatibility relationships more effectively. Higher AUC of our model shows it is better at distinguishing compatible vs. incompatible item pairs.

Table VI. shows improvements in evaluation metrics of our model compared to the HFGN baseline. Our proposed model achieves better performance on most metrics, especially on HR@10, Recall@10, and Precision@10, which are crucial for evaluating recommendation quality. Improvements are also observed in NDCG@10.

The model's accuracy improved by nearly 2%. These results indicate that concatenating image and text features enabled the extraction of more complex and accurate information from outfit items. Additionally, the use of multi-level graph attention mechanisms allowed dynamic consideration of each node's relative importance in the graph. Consequently, the proposed model's performance surpasses that of the baseline model.

TABLE IV. COMPARISON OF EXPERIMENTAL RESULTS OF THE PROPOSED MODEL WITH BASELINE MODELS FOR PERSONALIZED OUTFIT RECOMMENDATION.

| Model | NDCG@10 | Precision@10 | Recall@10 | HR@10 |
|---|---|---|---|---|
| FPITF | 0.0420 | 0.1121 | 0.0183 | 0.1006 |
| FHN | 0.0490 | 0.1192 | 0.0208 | 0.1109 |
| MF | 0.0872 | 0.2391 | 0.0434 | 0.2121 |
| VBPR | 0.0949 | 0.2481 | 0.0449 | 0.2201 |
| NGCF | 0.1143 | 0.3104 | 0.0554 | 0.2619 |
| HFGN | 0.1241 | 0.3390 | 0.0605 | 0.2833 |
| DTNM | 0.1813 | 0.1654 | 0.0625 | - |
| Try-On-CM | - | - | - | 0.2900 |
| RankBPR | - | - | **0.1921** | - |
| BCDSVD++ | 0.1666 | 0.1483 | 0.0569 | - |
| BPR | 0.2633 | 0.2410 | 0.0559 | - |
| LightGCN | 0.2882 | 0.2632 | 0.1015 | - |
| Hg-PDC | **0.3532** | 0.3080 | 0.1402 | - |
| **FGAT (ours)** | 0.1340 | **0.4424** | 0.1580 | **0.4286** |

TABLE V. COMPARISON OF ACCURACY AND AUC RESULTS OF THE PROPOSED MODEL WITH BASELINE MODELS FOR THE PREDICTING OUTFIT COMPATIBILITY SCORE AND MATCHING ITEMS IN FLTB TASK.

| Model | AUC | Accuracy |
|---|---|---|
| SiameseNet | 0.7087 | 0.5039 |
| Style2Vec | - | 0.6113 |
| Bi-LSTM | 0.7840 | 0.6384 |
| FOM | 0.8609 | 0.6879 |
| FHN | 0.8942 | 0.7422 |
| FaTrans-Multi | - | 0.776 |
| NGNN | 0.8381 | 0.8422 |
| HFGN | 0.875 | 0.8797 |
| **FGAT (ours)** | **0.8974** | **0.8956** |

TABLE VI. COMPARISON OF IMPROVEMENTS IN EVALUATION METRICS OF THE PROPOSED FGAT MODEL COMPARED TO THE BASELINE HFGN.

| Evaluation Metric | HFGN (Baseline) | FGAT (ours) | Improvement |
|---|---|---|---|
| HR@10 | 0.2833 | 0.4286 | 0.1453 |
| Recall@10 | 0.0605 | 0.1580 | 0.0975 |
| Precision@10 | 0.3390 | 0.4424 | 0.1034 |
| NDCG@10 | 0.1241 | 0.1340 | 0.0099 |
| Accuracy | 0.8797 | 0.8956 | 0.0159 |

## G. Personalized outfit Recommendation

To predict the probability of user u purchasing outfit o, we use the inner product of their final embeddings as follows:

$$\hat{y}_{uo} = {h_{u^*}}^T h_{o^*} \qquad (21)$$

This prediction equation transforms the prediction into an estimate of the similarity between user u and outfit o in a same hidden space.

The prediction score is a numerical value in the range [0, 1], where higher scores indicate greater similarity between the user's final embedding (representing their preferences) and the outfit. Based on these scores, the system recommends outfits to the user.

In Fig. 13, we present a three-level hierarchical graph for sample user 25. The black edges represent outfits previously purchased by the user, while the red dashed edges indicate the top 5 outfit recommendations—those that user 25 is most likely to purchase. This is modeled as a link prediction problem, and the model has predicted these links accordingly. User 25 has previously liked or purchased four outfits: 1378, 1426, 2069, and 4498. These outfits and their constituent items are shown in Fig. 14. The right side of the figure illustrates the newly recommended outfits for this user.

The output in Fig. 14 demonstrates that the FGAT model performs well in recommending new outfits to user 25. The recommendations are consistent with the user's preferences and previous purchases. Analysis of previous purchases shows that bag 378 appeared in two past purchases (outfits 375 and 4498) and is also present in the new recommended outfits. Likewise, hat 16951 was part of a previous purchase (outfit 4498) and appears again in the new suggestions.

The recommended outfit styles are determined based on user preferences, item compatibility, and the importance of co-occurring categories (e.g., shirt–pants or shirt–bag). By leveraging graph-based information propagation, message passing, and the concatenation of visual and textual features, this approach enhances recommendation precision. The results yield HR@10 of 0.4286, Precision@10 of 0.4424, and Recall@10 of 0.1580.

*H. predicting outfit compatibility score*

To determine whether a set of items forms a compatible and appropriate outfit, we use the item representation to calculate a compatibility matching score. Unlike existing methods that simply add up pairwise matching scores, we believe that each item in an outfit has different importance. For example, in an outfit, maybe a long dress defines the overall style and is more important than accessories. To determine the importance of items in an outfit, we use a self-attention mechanism that creates a R-view attention and a R-view compatibility score. Our model attempts to assess how well a set of fashion items are matched. It examines compatibility not only by examining items pairwise, but also by considering the overall importance and contribution of each item to an outfit set. In the proposed model, we set the number of views (R) to 6.

❖ R-View Attention

The R-view attention mapp learns how important each item is in each semantic view (e.g., style, color coordination, brand coherence) within an outfit. It is computed as follow Equation (22). $O_{em} \in R^{n \times d}$ is Embedding matrix of an outfit with n items and embedding dimension d, and $W_5 \in R^{v \times d}$, $W_4 \in R^{R \times v}$: Learnable weight matrices. Softmax function, applied row-wise to normalize attention scores.:

$$A_{rm} = Softmax(W_4 leakyRelu(W_5 O_{em}^T)) \qquad 22$$

Example: Given 3 items (shirt, jeans, shoes) with 64-dimensional embeddings, the attention matrix might look like the Table VII. This matrix shows how much attention should be given to each item in each semantic view.

❖ R-View Compatibility Score

After attention is computed, the compatibility score mapp learns how compatible each item is with the rest of the outfit under each view:

$$C_{rm} = tanh(W_6 \, leakyRelu(W_7 O_{em}^T)) \qquad 23$$

These scores reflect how well each item compatible with others in the outfit under each view (Table VIII.)

❖ Weighted Outfit Compatibility Score

The final weighted compatibility score for the outfit is computed by combining attention weights and compatibility scores:

$$\hat{s}_o = \sum_{r=1}^{R} a_r^T c_r \qquad 24$$

Where $ar$ and $cr$ are the r-th rows from attention matrix $A$ and compatibility matrix $C$, respectively.

Example Calculation:

View 1 (Style): a = [0.6, 0.2, 0.2], c = [0.8, 0.5, 0.2] → 0.62

View 2 (Color): a = [0.3, 0.5, 0.2], c = [0.4, 0.7, 0.9] → 0.65

View 3 (Brand): a = [0.1, 0.3, 0.6], c = [0.3, 0.5, 0.3] → 0.36

Final outfit compatibility score = (0.62 + 0.65 + 0.36) / 3 =0.5433.

❖ FLTB Task

To evaluate the model's performance on outfit compatibility, the FLTB task is used. One item in a test outfit is masked. Three items are randomly selected from other outfits. The model scores outfit by all four candidates. The item with the highest compatibility score is selected as the best match. Fig. 15 shows an example of FLTB.

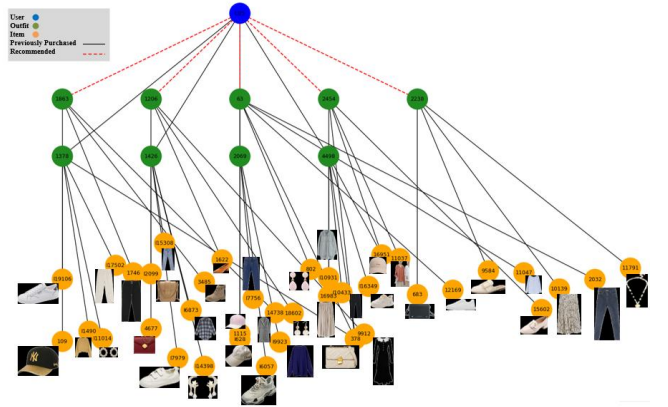

Fig. 13. Three-level hierarchical graph for recommending the top 5 outfits to user 25 - Link prediction.

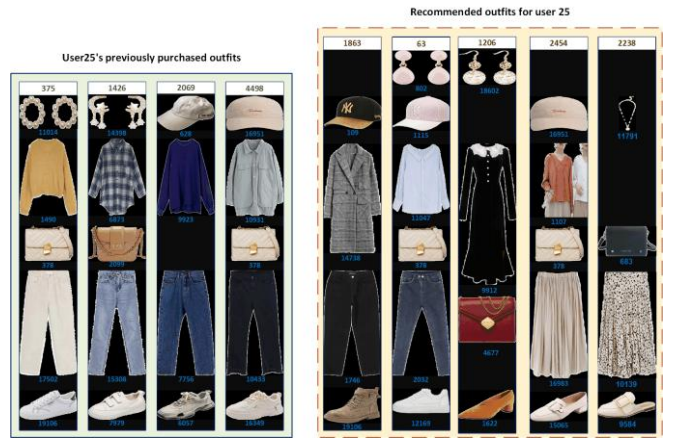

Fig. 14. Previously purchased outfits of user 25 and the model's recommended outfits.

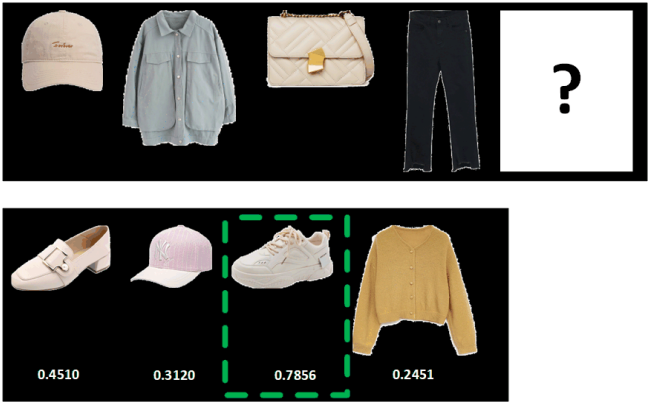

Fig. 15. Example of the FLTB task. The model selects the item with the highest compatibility score as the complementary item for the outfit.

TABLE VII. Example of an R-view Attention Matrix. Each Row Represents a View, and Each Column Shows the Importance of an Item in That View.

| View | Shirt | Jeans | Shoes |
|---|---|---|---|
| *Style* | 0.6 | 0.2 | 0.2 |
| *Color* | 0.3 | 0.5 | 0.2 |
| *Brand* | 0.1 | 0.3 | 0.6 |

TABLE VIII. Example of an R-view Compatibility Score Matrix. Each Row Represents a View, and Each Column Represents a Compatibility Score of Each Item With Other Outfit Items in That View.

| View | Shirt | Jeans | Shoes |
|---|---|---|---|
| *Style* | 0.8 | 0.5 | 0.2 |
| *Color* | 0.4 | 0.7 | 0.9 |
| *Brand* | 0.3 | 0.5 | 0.3 |

## V. CONCLUSION

This research proposed and implemented FGAT, a hierarchical graph-based deep learning model for personalized and compatibility-aware outfit recommendation. Addressing limitations in previous models, such as their inability to simultaneously capture user preferences and item compatibility, the framework effectively models complex relationships among users, outfits, and items through a three-level hierarchical fashion graph and multi-head attention mechanisms. By integrating visual and textual features with semantic representations, FGAT enhances the perception of outfit compatibility. Its architecture allows dynamic weighting of nodes, leading to superior performance in both recommendation precision and compatibility matching. Experimental results on the POG dataset confirmed the model's superiority over baselines such as HFGN across multiple evaluation metrics, highlighting the effectiveness of a graph-based, multi-modal, and attention-driven design for fashion recommendation.

From an application perspective, recommender systems (RS) play a critical role in modern e-commerce, especially in fashion, where personalization directly affects user engagement, satisfaction, and sales conversion. Beyond improving individual outfit suggestions, RS can enhance the overall shopping experience by reducing decision fatigue, promoting style discovery, and enabling adaptive personalization that mirrors real-world fashion advisors. Thus, advances in recommendation models like FGAT not only contribute academically but also hold substantial commercial and practical importance in shaping next-generation intelligent shopping platforms.

Despite its promising results, this research also faces several challenges. The first limitation lies in the static nature of the dataset, which does not account for evolving user preferences over time. Additionally, the computational complexity of hierarchical graph neural networks presents scalability issues, particularly when training on large-scale fashion graphs without access to high-performance GPU clusters. Another challenge is the data imbalance across categories, where items with fewer samples make it harder for the model to generalize. Furthermore, while FGAT primarily focuses on first-order and local relationships, higher-order and long-range dependencies remain underexplored due to resource constraints.

To further improve the model, future work can address these challenges in several directions:

- Incorporating time-aware and dynamic user-item interactions, enabling the model to adapt to evolving preferences.

- Incorporating time-aware and dynamic user-item interactions, enabling the model to adapt to evolving preferences.

- Integrating contextual signals such as season, weather, occasions, and current fashion trends to provide adaptive and situational recommendations.

- Leveraging Transformer-based architectures (e.g., multimodal transformers) to capture deeper sequential, semantic, and multimodal dependencies, potentially improving both personalization and interpretability.

REFERENCES

[1] F. T. Abdul Hussien, A. M. S. Rahma, and H. B. Abdul Wahab, "Recommendation Systems For E-commerce Systems An Overview," J. Phys. Conf. Ser., vol. 1897, no. 1, p. 012024, May 2021, doi: 10.1088/1742-6596/1897/1/012024.

[2] A. Veit, B. Kovacs, S. Bell, J. McAuley, K. Bala, and S. Belongie, "Learning Visual Clothing Style with Heterogeneous Dyadic Co-Occurrences," in 2015 IEEE International Conference on Computer Vision (ICCV), Santiago, Chile: IEEE, Dec. 2015, pp. 4642–4650. doi: 10.1109/ICCV.2015.527.

[3] W. Guan, H. Wen, X. Song, C.-H. Yeh, X. Chang, and L. Nie, "Multimodal Compatibility Modeling via Exploring the Consistent and Complementary Correlations," in Proceedings of the 29th ACM International Conference on Multimedia, Virtual Event China: ACM, Oct. 2021, pp. 2299–2307. doi: 10.1145/3474085.3475392.

[4] A. Gershman, A. Meisels, K.-H. Lüke, L. Rokach, A. Schclar, and A. Sturm, "A decision tree based recommender system," presented at the 10th International Conferenceon Innovative Internet Community Systems (I2CS) – Jubilee Edition 2010 –, Gesellschaft für Informatik e.V., 2010, pp. 170–179. Accessed: Aug. 02, 2024. [Online]. Available: https://dl.gi.de/handle/20.500.12116/19012

[5] T. N. Kipf and M. Welling, "Semi-Supervised Classification with Graph Convolutional Networks," Feb. 22, 2017, arXiv: arXiv:1609.02907. doi: 10.48550/arXiv.1609.02907.

[6] W. Hamilton, Z. Ying, and J. Leskovec, "Inductive Representation Learning on Large Graphs," in Advances in Neural Information Processing Systems, Curran Associates, Inc., 2017. Accessed: Aug.

[7] X. Han, L. Yu, X. Zhu, L. Zhang, Y.-Z. Song, and T. Xiang, "FashionViL: Fashion-Focused Vision-and-Language Representation Learning," in Computer Vision – ECCV 2022, S. Avidan, G. Brostow, M. Cissé, G. M. Farinella, and T. Hassner, Eds., Cham: Springer Nature Switzerland, 2022, pp. 634–651. doi: 10.1007/978-3-031-19833-5_37.

[8] S. Cao, W. Lu, and Q. Xu, "GraRep: Learning Graph Representations with Global Structural Information," in Proceedings of the 24th ACM International on Conference on Information and Knowledge Management, Melbourne Australia: ACM, Oct. 2015, pp. 891–900. doi: 10.1145/2806416.2806512.

[9] R. Fontenot, J. Lazarus, P. Rudick, and A. Sgambellone, "Hierarchical Neural Networks (HNN): Using TensorFlow to build HNN," SMU Data Sci. Rev., vol. 6, no. 2, Sep. 2022, [Online]. Available: https://scholar.smu.edu/datasciencereview/vol6/iss2/4

[10] X. Li, X. Wang, X. He, L. Chen, J. Xiao, and T.-S. Chua, "Hierarchical Fashion Graph Network for Personalized Outfit Recommendation," May 26, 2020, arXiv: arXiv:2005.12566. Accessed: Feb. 01, 2024. [Online]. Available: http://arxiv.org/abs/2005.12566

[11] Z. Cui, Z. Li, S. Wu, X.-Y. Zhang, and L. Wang, "Dressing as a Whole: Outfit Compatibility Learning Based on Node-wise Graph Neural Networks," in The World Wide Web Conference, in WWW '19. New York, NY, USA: Association for Computing Machinery, May 2019, pp. 307–317. doi: 10.1145/3308558.3313444.

[12] Z. Lu, Y. Hu, Y. Jiang, Y. Chen, and B. Zeng, "Learning Binary Code for Personalized Fashion Recommendation," presented at the Proceedings of the IEEE/CVF Conference on Computer Vision and Pattern Recognition, 2019, pp. 10562–10570. Accessed: Aug. 19, 2024. [Online]. Available: https://openaccess.thecvf.com/content_CVPR_2019/html/Lu_Learning_Binary_Code_for_Personalized_Fashion_Recommendation_CVPR_2019_paper.html

[13] D. Marcus, Graph Theory, vol. 53. in AMS/MAA Textbooks, vol. 53. Providence, Rhode Island: American Mathematical Society, 2020. doi: 10.1090/text/053.

[14] K. Kalashi, S. Saed, and B. Teimourpour, "Sentiment-Driven Community Detection in a Network of Perfume Preferences," Oct. 28, 2024, arXiv: arXiv:2410.19177. doi: 10.48550/arXiv.2410.19177.

[15] V. T. Hoang, H.-J. Jeon, E.-S. You, Y. Yoon, S. Jung, and O.-J. Lee, "Graph Representation Learning and Its Applications: A Survey," Sensors, vol. 23, no. 8, Art. no. 8, Jan. 2023, doi: 10.3390/s23084168.

[16] B. Perozzi, R. Al-Rfou, and S. Skiena, "DeepWalk: Online Learning of Social Representations," in Proceedings of the 20th ACM SIGKDD international conference on Knowledge discovery and data mining, Aug. 2014, pp. 701–710. doi: 10.1145/2623330.2623732.

[17] M. Ou, P. Cui, J. Pei, Z. Zhang, and W. Zhu, "Asymmetric Transitivity Preserving Graph Embedding," in Proceedings of the 22nd ACM SIGKDD International Conference on Knowledge Discovery and Data Mining, San Francisco California USA: ACM, Aug. 2016, pp. 1105–1114. doi: 10.1145/2939672.2939751.

[18] M. A. Soltanshahi, B. Teimourpour, T. Khatibi, and H. Zare, "GrAR: A novel framework for Graph Alignment based on Relativity concept," Expert Syst. Appl., vol. 187, p. 115908, Jan. 2022, doi: 10.1016/j.eswa.2021.115908.

[19] Z. Wu, S. Pan, F. Chen, G. Long, C. Zhang, and P. S. Yu, "A Comprehensive Survey on Graph Neural Networks," IEEE Trans. Neural Netw. Learn. Syst., vol. 32, no. 1, pp. 4–24, Jan. 2021, doi: 10.1109/TNNLS.2020.2978386.

[20] X. Wang, X. He, M. Wang, F. Feng, and T.-S. Chua, "Neural Graph Collaborative Filtering," in Proceedings of the 42nd International ACM SIGIR Conference on Research and Development in Information Retrieval, Jul. 2019, pp. 165–174. doi: 10.1145/3331184.3331267.

[21] P. Veličković, G. Cucurull, A. Casanova, A. Romero, P. Liò, and Y. Bengio, "Graph Attention Networks," Feb. 04, 2018, arXiv: arXiv:1710.10903. Accessed: Aug. 01, 2024. [Online]. Available: http://arxiv.org/abs/1710.10903

[22] Z. Zhou, View Profile, Z. Su, View Profile, R. Wang, and View Profile, "Attribute-aware heterogeneous graph network for fashion compatibility prediction," Neurocomputing, vol. 495, no. C, pp. 62–74, Jul. 2022, doi: 10.1016/j.neucom.2022.04.121.

[23] S. Chakraborty, Md. S. Hoque, N. Rahman Jeem, M. C. Biswas, D. Bardhan, and E. Lobaton, "Fashion Recommendation Systems, Models and Methods: A Review," Informatics, vol. 8, no. 3, p. 49, Jul. 2021, doi: 10.3390/informatics8030049.

[24] S. Gulati, "Fashion Recommendation: Outfit Compatibility using GNN," Apr. 28, 2024, arXiv: arXiv:2404.18040. doi: 10.48550/arXiv.2404.18040.

[25] Y. Ding, Y. Ma, W. K. Wong, and T.-S. Chua, "Leveraging Two Types of Global Graph for Sequential Fashion Recommendation," in Proceedings of the 2021 International Conference on Multimedia Retrieval, Aug. 2021, pp. 73–81. doi: 10.1145/3460426.3463638.

[26] C. Li, I. Ishak, H. Ibrahim, M. Zolkepli, F. Sidi, and C. Li, "Deep Learning-Based Recommendation System: Systematic Review and Classification," IEEE Access, vol. 11, pp. 113790–113835, 2023, doi: 10.1109/ACCESS.2023.3323353.

[27] H. Liu, L. Li, N. Yu, K. Ma, T. Peng, and X. Hu, "Outfit compatibility model using fully connected self-adjusting graph neural network," Vis. Comput., vol. 40, no. 11, pp. 8331–8343, Nov. 2024, doi: 10.1007/s00371-023-03238-6.

[28] F. Chen, Y.-C. Wang, B. Wang, and C.-C. J. Kuo, "Graph representation learning: a survey," APSIPA Trans. Signal Inf. Process., vol. 9, no. 1, 2020, doi: 10.1017/ATSIP.2020.13.

[29] H. Zhang, X. Yang, J. Tan, C.-H. Wu, J. Wang, and C.-C. J. Kuo, "Learning Color Compatibility in Fashion Outfits," Jul. 05, 2020, arXiv: arXiv:2007.02388. doi: 10.48550/arXiv.2007.02388.

[30] J. Xu, Y. Wei, A. Wang, H. Zhao, and D. Lefloch, "Analysis of Clothing Image Classification Models: A Comparison Study between Traditional Machine Learning and Deep Learning Models," Fibres Text. East. Eur., vol. 30, no. 5, pp. 66–78, Oct. 2022, doi: 10.2478/ftee-2022-0046.

[31] S. Saed, B. Teimourpour, K. Kalashi, and M. A. Soltanshahi, "An Efficient Multiple Convolutional Neural Network Model (MCNN-14) for Fashion Image Classification," in 2024 10th International Conference on Web Research (ICWR), Apr. 2024, pp. 13–21. doi: 10.1109/ICWR61162.2024.10533341.

[32] A. Schindler, T. Lidy, S. Karner, and M. Hecker, "Fashion and Apparel Classification using Convolutional Neural Networks," Nov. 11, 2018, arXiv: arXiv:1811.04374. doi: 10.48550/arXiv.1811.04374.

[33] A. Krizhevsky, I. Sutskever, and G. E. Hinton, "ImageNet Classification with Deep Convolutional Neural Networks," in Advances in Neural Information Processing Systems, Curran Associates, Inc., 2012. Accessed: Aug. 17, 2024. [Online]. Available: https://proceedings.neurips.cc/paper/2012/hash/c399862d3b9d6b76c8436e924a68c45b-Abstract.html

[34] K. He, X. Zhang, S. Ren, and J. Sun, "Deep Residual Learning for Image Recognition," presented at the Proceedings of the IEEE Conference on Computer Vision and Pattern Recognition, 2016, pp. 770–778. Accessed: Aug. 17, 2024. [Online]. Available: https://openaccess.thecvf.com/content_cvpr_2016/html/He_Deep_Residual_Learning_CVPR_2016_paper.html

[35] Z. Liu, P. Luo, S. Qiu, X. Wang, and X. Tang, "DeepFashion: Powering Robust Clothes Recognition and Retrieval With Rich Annotations," presented at the Proceedings of the IEEE Conference on Computer Vision and Pattern Recognition, 2016, pp. 1096–1104. Accessed: Aug. 17, 2024. [Online]. Available: https://openaccess.thecvf.com/content_cvpr_2016/html/Liu_DeepFashion_Powering_Robust_CVPR_2016_paper.html

[36] MaJianghong, SunHuiyue, YangDezhao, and ZhangHaijun, "Personalized Fashion Recommendations for Diverse Body Shapes with Contrastive Multimodal Cross-Attention Network," ACM Trans. Intell. Syst. Technol., Jul. 2024, doi: 10.1145/3637217.

[37] R. Xu, J. Wang, and Y. Li, "Heterogeneous-Grained Multi-Modal Graph Network for Outfit Recommendation," IEEE Trans. Emerg. Top. Comput. Intell., vol. 8, no. 2, pp. 1788–1799, Apr. 2024, doi: 10.1109/TETCI.2024.3358190.

[38] S. Shirkhani, H. Mokayed, R. Saini, and H. Y. Chai, "Study of AI-Driven Fashion Recommender Systems," SN Comput. Sci., vol. 4, no. 5, p. 514, Jul. 2023, doi: 10.1007/s42979-023-01932-9.

[39] M. I. Vasileva, B. A. Plummer, K. Dusad, S. Rajpal, R. Kumar, and D. Forsyth, "Learning Type-Aware Embeddings for Fashion Compatibility," presented at the European Conference on Computer Vision (ECCV), 2018, pp. 390–405. Accessed: Aug. 17, 2024. [Online]. Available: https://openaccess.thecvf.com/content_ECCV_2018/html/Mariya_Vasileva_Learning_Type-Aware_Embeddings_ECCV_2018_paper.html

[40] W. Zong, "Dress Style Recommendation Based on Female Body Shapes," Aug. 2022, Accessed: Aug. 17, 2024. [Online]. Available: https://hdl.handle.net/1813/112171

[41] S. Zhao et al., "Affective Image Content Analysis: Two Decades Review and New Perspectives," IEEE Trans. Pattern Anal. Mach. Intell., vol. 44, no. 10, pp. 6729–6751, Oct. 2022, doi: 10.1109/TPAMI.2021.3094362.

[42] A. Felfernig et al., "Recommender systems for sustainability: overview and research issues," Front. Big Data, vol. 6, Oct. 2023, doi: 10.3389/fdata.2023.1284511.

[43] G. Salton and C. Buckley, "Term-weighting approaches in automatic text retrieval," Inf. Process. Manag., vol. 24, no. 5, pp. 513–523, Jan. 1988, doi: 10.1016/0306-4573(88)90021-0.

[44] T. Mikolov, K. Chen, G. Corrado, and J. Dean, "Efficient Estimation of Word Representations in Vector Space," Sep. 06, 2013, arXiv: arXiv:1301.3781. doi: 10.48550/arXiv.1301.3781.

[45] A. Vaswani et al., "Attention Is All You Need," Aug. 01, 2023, arXiv: arXiv:1706.03762. doi: 10.48550/arXiv.1706.03762.

[46] J. Devlin, M.-W. Chang, K. Lee, and K. Toutanova, "BERT: Pre-training of Deep Bidirectional Transformers for Language Understanding," May 24, 2019, arXiv: arXiv:1810.04805. doi: 10.48550/arXiv.1810.04805.

[47] Y. Du, Z. Liu, J. Li, and W. X. Zhao, "A Survey of Vision-Language Pre-Trained Models," Jul. 15, 2022, arXiv: arXiv:2202.10936. doi: 10.48550/arXiv.2202.10936.

[48] F. Sun et al., "BERT4Rec: Sequential Recommendation with Bidirectional Encoder Representations from Transformer," in Proceedings of the 28th ACM International Conference on Information and Knowledge Management, Beijing China: ACM, Nov. 2019, pp. 1441–1450. doi: 10.1145/3357384.3357895.

[49] T. Hofmann, "Probabilistic Latent Semantic Analysis," Jan. 23, 2013, arXiv: arXiv:1301.6705. doi: 10.48550/arXiv.1301.6705.

[50] X. He, L. Liao, H. Zhang, L. Nie, X. Hu, and T.-S. Chua, "Neural Collaborative Filtering," in Proceedings of the 26th International Conference on World Wide Web, in WWW '17. Republic and Canton of Geneva, CHE: International World Wide Web Conferences Steering Committee, Apr. 2017, pp. 173–182. doi: 10.1145/3038912.3052569.

[51] G. Cucurull, P. Taslakian, and D. Vazquez, "Context-Aware Visual Compatibility Prediction," presented at the Proceedings of the IEEE/CVF Conference on Computer Vision and Pattern Recognition, 2019, pp. 12617–12626. Accessed: Aug. 19, 2024. [Online]. Available: https://openaccess.thecvf.com/content_CVPR_2019/html/Cucurull_Context-Aware_Visual_Compatibility_Prediction_CVPR_2019_paper.html

[52] H. Lee, J. Seol, and S. Lee, "Style2Vec: Representation Learning for Fashion Items from Style Sets," Aug. 14, 2017, arXiv: arXiv:1708.04014. doi: 10.48550/arXiv.1708.04014.

[53] R. Sarkar et al., "OutfitTransformer: Learning Outfit Representations for Fashion Recommendation," in 2023 IEEE/CVF Winter Conference on Applications of Computer Vision (WACV), Waikoloa, HI, USA: IEEE, Jan. 2023, pp. 3590–3598. doi: 10.1109/WACV56688.2023.00359.

[54] W. Chen et al., "POG: Personalized Outfit Generation for Fashion Recommendation at Alibaba iFashion," in Proceedings of the 25th ACM SIGKDD International Conference on Knowledge Discovery & Data Mining, in KDD '19. New York, NY, USA: Association for Computing Machinery, Jul. 2019, pp. 2662–2670. doi: 10.1145/3292500.3330652.

[55] W. Guan et al., "Partially Supervised Compatibility Modeling," IEEE Trans. Image Process., vol. 31, pp. 4733–4745, 2022, doi: 10.1109/TIP.2022.3187290.

[56] S. Abu-El-Haija, A. Kapoor, B. Perozzi, and J. Lee, "N-GCN: Multi-scale Graph Convolution for Semi-supervised Node Classification," in Proceedings of The 35th Uncertainty in Artificial Intelligence Conference, PMLR, Aug. 2020, pp. 841–851. Accessed: Aug. 19, 2024. [Online]. Available: https://proceedings.mlr.press/v115/abu-el-haija20a.html

[57] L. Hu et al., "Graph Neural News Recommendation with Unsupervised Preference Disentanglement," in Proceedings of the 58th Annual Meeting of the Association for Computational Linguistics, D. Jurafsky, J. Chai, N. Schluter, and J. Tetreault, Eds., Online: Association for Computational Linguistics, Jul. 2020, pp. 4255–4264. doi: 10.18653/v1/2020.acl-main.392.

[58] W. Guan, X. Song, H. Zhang, M. Liu, C.-H. Yeh, and X. Chang, "Bi-directional Heterogeneous Graph Hashing towards Efficient Outfit Recommendation," in Proceedings of the 30th ACM International Conference on Multimedia, in MM '22. New York, NY, USA: Association for Computing Machinery, Oct. 2022, pp. 268–276. doi: 10.1145/3503161.3548020.

[59] S. Gulati, "Fashion Recommendation: Outfit Compatibility using GNN," Apr. 28, 2024, arXiv: arXiv:2404.18040. doi: 10.48550/arXiv.2404.18040.

[60] Mingming L., Qingming G., Ya Z., and Xiankang Y., "Heterogeneous Graph Neural Network for Fashion Outfit Compatibility Prediction," J. Comput.-Aided Des. Comput. Graph., vol. 36, no. 9, pp. 1351–1361, 2024, doi: 10.3724/SP.J.1089.2024.19974.

[61] R. He and J. McAuley, "VBPR: Visual Bayesian Personalized Ranking from Implicit Feedback," Proc. AAAI Conf. Artif. Intell., vol. 30, no. 1, Art. no. 1, Feb. 2016, doi: 10.1609/aaai.v30i1.9973.

[62] D. Sagar, J. Garg, P. Kansal, S. Bhalla, R. R. Shah, and Y. Yu, "PAI-BPR: Personalized Outfit Recommendation Scheme with Attribute-wise Interpretability," in 2020 IEEE Sixth International Conference on Multimedia Big Data (BigMM), Sep. 2020, pp. 221–230. doi: 10.1109/BigMM50055.2020.00039.

[63] X. Dong, J. Wu, X. Song, H. Dai, and L. Nie, "Fashion Compatibility Modeling through a Multi-modal Try-on-guided Scheme," in Proceedings of the 43rd International ACM SIGIR Conference on Research and Development in Information Retrieval, in SIGIR '20. New York, NY, USA: Association for Computing Machinery, Jul. 2020, pp. 771–780. doi: 10.1145/3397271.3401047.

[64] X. Song, S.-T. Fang, X. Chen, Y. Wei, Z. Zhao, and L. Nie, "Modality-Oriented Graph Learning Toward Outfit Compatibility Modeling," IEEE Trans. Multimed., vol. 25, pp. 856–867, 2023, doi: 10.1109/TMM.2021.3134164.

[65] Y. Hu, X. Yi, and L. S. Davis, "Collaborative Fashion Recommendation: A Functional Tensor Factorization Approach," in Proceedings of the 23rd ACM international conference on Multimedia, in MM '15. New York, NY, USA: Association for Computing Machinery, Oct. 2015, pp. 129–138. doi: 10.1145/2733373.2806239.

[66] X. Song, X. Han, Y. Li, J. Chen, X.-S. Xu, and L. Nie, "GP-BPR: Personalized Compatibility Modeling for Clothing Matching," in Proceedings of the 27th ACM International Conference on Multimedia, in MM '19. New York, NY, USA: Association for Computing Machinery, Oct. 2019, pp. 320–328. doi: 10.1145/3343031.3350956.

[67] W. Guan, F. Jiao, X. Song, H. Wen, C.-H. Yeh, and X. Chang, "Personalized Fashion Compatibility Modeling via Metapath-guided Heterogeneous Graph Learning," in Proceedings of the 45th International ACM SIGIR Conference on Research and Development in Information Retrieval, in SIGIR '22. New York, NY, USA: Association for Computing Machinery, Jul. 2022, pp. 482–491. doi: 10.1145/3477495.3532038.

[68] B. S. Vivek, G. Bhattacharya, J. Gubbi, B. L. V, A. Pal, and P. Balamuralidhar, "Personalized Outfit Compatibility Prediction Using Outfit Graph Network," in 2023 International Joint Conference on Neural Networks (IJCNN), Gold Coast, Australia: IEEE, Jun. 2023, pp. 1–8. doi: 10.1109/IJCNN54540.2023.10191458.

[69] Y. Wang, L. Liu, X. Fu, and L. Liu, "MCCP: multi-modal fashion compatibility and conditional preference model for personalized clothing recommendation," Multimed. Tools Appl., vol. 83, no. 4, pp. 9621–9645, Jan. 2024, doi: 10.1007/s11042-023-15659-5.

[70] R. Xu, J. Wang, M. Li, J. Xu, and Y. Li, "Hierarchical Multimodal Graph Learning for Outfit Compatibility Modelling," IEEE Trans. Emerg. Top. Comput. Intell., vol. 8, no. 6, pp. 4130–4142, Dec. 2024, doi: 10.1109/TETCI.2024.3386774.

[71] X. Su et al., "Personalized clothing recommendation fusing the 4-season color system and users' biological characteristics," Multimed. Tools Appl., vol. 83, no. 5, pp. 12597–12625, Feb. 2024, doi: 10.1007/s11042-023-16014-4.

[72] A. Shoeb, M. H. Ali, Md. M. Ali, and M. S. Qaseem, "PFRS: Personalized Fashion Recommendation System Using EfficientNet," in 2024 International Conference on Innovation and Intelligence for Informatics, Computing, and Technologies (3ICT), Nov. 2024, pp. 337–344. doi: 10.1109/3ict64318.2024.10824272.

[73] M. C. Jung, J. Monteil, P. Schulz, and V. Vaskovych, "Personalised Outfit Recommendation via History-aware Transformers," Sep. 27, 2024, arXiv: arXiv:2407.00289. doi: 10.48550/arXiv.2407.00289.

[74] Y. Dang, Z. Pan, X. Zhang, W. Chen, F. Cai, and H. Chen, "Discrepancy Learning Guided Hierarchical Fusion Network for Multi-modal Recommendation," Knowl.-Based Syst., vol. 317, p. 113496, May 2025, doi: 10.1016/j.knosys.2025.113496.

[75] Y.-H. Hu, T.-H. Liu, K. Chen, and F.-C. Yeh, "Leveraging meta-path and co-attention to model consumer preference stability in fashion recommendations," Decis. Support Syst., vol. 194, p. 114455, Jul. 2025, doi: 10.1016/j.dss.2025.114455.

[76] Y. Bi, Y. Gao, and H. Wang, "StePO-Rec: Towards Personalized Outfit Styling Assistant via Knowledge-Guided Multi-Step Reasoning," Apr. 14, 2025, arXiv: arXiv:2504.09915. doi: 10.48550/arXiv.2504.09915.

[77] W. Guan, X. Song, D. Zhou, and L. Nie, "Hashing-Based Efficient Outfit Recommendation," in Advanced Multimodal Compatibility Modeling and Recommendation, W. Guan, X. Song, D. Zhou, and L. Nie, Eds., Cham: Springer Nature Switzerland, 2025, pp. 103–122. doi: 10.1007/978-3-031-81048-0_6.

[78] C. Chen, F. Mo, X. Fan, and H. Yamana, "Personalized Fashion Recommendation with Image Attributes and Aesthetics Assessment," Jan. 06, 2025, arXiv: arXiv:2501.03085. doi: 10.48550/arXiv.2501.03085.

[79] S. Rendle, C. Freudenthaler, Z. Gantner, and L. Schmidt-Thieme, "BPR: Bayesian Personalized Ranking from Implicit Feedback," May 09, 2012, arXiv: arXiv:1205.2618. doi: 10.48550/arXiv.1205.2618.